# Solar Cell Surface Defect Inspection Based on Multispectral Convolutional Neural Network

Haiyong Chen[1,2], Yue Pang[1], Qidi Hu[1], Kun Liu[1,2]

**Abstract** Similar and indeterminate defect detection of solar cell surface with heterogeneous texture and complex background is a challenge of solar cell manufacturing. The traditional manufacturing process relies on human eye detection which requires a large number of workers without a stable and good detection effect. In order to solve the problem, a visual defect detection method based on multi-spectral deep convolutional neural network is designed in this paper. Firstly, a selected convolutional neural network （CNN）model is established. By adjusting the depth and width of the model, the influence of model depth and kernel size on the recognition result is evaluated. The optimal convolutional neural network model structure is selected. Secondly, the light spectrum features of solar cell color image are analyzed. It is found that a variety of defects exhibited different distinguishable characteristics in different spectral bands. Thus, a multi-spectral convolutional neural network model is constructed to enhance the discrimination ability of the model to distinguish between complex texture background features and defect features. Finally, some experimental results and K-fold cross validation show that the multi-spectral deep convolutional neural network model can effectively detect the solar cell surface defects with higher accuracy and greater adaptability. The accuracy of defect recognition reaches 94.30%. Applying such an algorithm can increase the efficiency of solar cell manufacturing and make the manufacturing process smarter.

## 1 Introduction

As one of the most important renewable energy sources, solar energy is gaining more and more attention. However, in the manufacturing process, solar cells will have some surface defects, including broken gates, pasting spot, thick lines, dirty cells, missing corners, scratches, chromatic aberrations, etc. Solar cells with defects should be detected and eliminated in time to avoid the quality damage of solar cell module in the next step of production. Therefore, surface defect detection of solar cells plays a key role in controlling the quality of solar cell products during manufacturing process [1] .

As machine vision develops rapidly, an image-based defect detection method has been employed for solar cell surface quality controlling in manufacturing industry. Solar cell surface quality inspection can not only improve the production quality of the solar cell module, but also increase the lifetime of the solar cell module. Generally, solar cells are divided into monocrystalline silicon and polysilicon by the production materials. The monocrystalline silicon solar cell has a uniform background texture.

The defect detection object of this paper is polycrystalline silicon solar cells. The surface of polycrystalline silicon solar cells contains a large number of lattice particles with random shapes and sizes, which are randomly distributed in different directions and locations. Moreover, the color features of surface defects in such complex background vary randomly. Thus, the non-uniform backgrounds and complex textures bring a huge challenge for visual inspection of multiple defects of solar cell.

## 2 Related works on solar cell surface detection

In order to reliably obtain the surface defect characteristics, some feature extraction methods are effective when image intensity consistency is satisfied. The existing surface defect detection methods based on machine vision can be classified into four categories in term of texture surface features: 1) non-textured surface; 2) repeated pattern surface; 3) homogeneously-textured surface;4) non-homogeneously-textured surface. For non-textured surface images, such as sheet steel [2][3], glass screen [5] or integrated circuit[5] ,the commonly used texture features are statistical measures[6], for instance first-order statistics (i.e., mean and variance) and second-order statistics [7]. For repeated pattern surface images, such as textile fabrics[8]**.**, semiconductor wafers [9]**.**. The detection algorithms usually use template matching methods between current image and self-generating template. For homogeneously-textured surface images, the texture pattern shows high similarity everywhere in the image, such as wood inspection[10], there are two kinds of defect detection method, spatial and spectral approaches. In the spatial domain, defect can be effectively identified with statistics features derived from co-occurrence matrices [11]. In the spectral domain, PJR Torres et.al[12] proposes an algorithm for checking thermal fuses with machine vision to detect four different defects. For non-homogeneously-textured surface images, such as marble or granite, the Ref [13]**.** constructed a feature extraction system for marble tile inspection by employing eight Gabor filters. Liu et.al [14] proposed a texture edge detection method that includes encoding and prediction modules for texture inspection. Mirmehdizai [15] presented an automatic defect detection method for random color texture surface. However, the detection of surface defects of solar cells is a multi-feature extraction and detection problem under a non-uniform texture background. The polycrystalline solar cell always shows the complex surface with inhomogeneous texture and non-Gaussian color distribution. One of typical features of these defects is that they have different characteristics in different spectral ranges.

Many existing solar cell defect detection methods focus on the analysis of electroluminescence (EL) infrared images under 1000nm-1200nm wave length. Chiou et al.[16] developed a regional growth detection algorithm to extract cracks defect

✉ Kun Liu
   liukun@hebut.edu.cn
1  The School of Artificial Intelligence, Hebei University of Technology, Tianjin 300130, China
2  Intelligent Rehabilitation Equipment and Detection Technology Engineering Research Center of Ministry of Education, Tianjin 300130, China from the captured images. Fu et al. [17] proposed a method for detecting cracks in solar cells using machine vision. This method can only identify defects on the edge of the cell and has no significant effect on internal defects. Anwar [18] proposed an improved anisotropic diffusion filter and image segmentation algorithm for the detection of micro-cracks in polycrystalline silicon solar cells with a detection result of 88%. This method is only applicable to micro-crack detection and cannot detect multiple defects. Tsai et al. [19] proposed a method based on independent component analysis to evaluate the reconstruction error between the detected image and the reconstructed image to detect the defects. This method can cause erroneous detection of defects in non-uniform light areas. Tsai also proposed a clustering algorithm for solar cell surface defect detection. The algorithm uses binary tree clustering algorithm to cluster the distribution of multiple sets of training data and determines the defect type by calculating the distance between classes. Ordaz [20] used the gray distribution histogram in the EL image of the cell for analysis. However, this method can only obtain the gray region distribution and statistical information of the image. It is unable to extract and distinguish the texture features, and it lacks the ability to recognize local small defects. Unfortunately, this method only has significant effect on linear features and performs poorly on other defections in the image with visible light spectrum. Qian et al. [21]. reviewed the typical types of solar cell surface defects and evaluated current popular machine vision detection algorithms. For the four types of defects, the accuracy rate on the test data set reached about 95%. However, the features depend on manual selection and the number of experimental samples is small. The above shortcomings restrict the adaptability. Li et al. [22] proposed a discriminant method based on wavelet transform for the detection of defects in polycrystalline silicon solar cells. The experimental results show that the method has good effects on fingerprints, dirty marks, etc. However, the effect of sharp edges on the edges is poor and it cannot be applied to all defects. Yao et al. [23] used Robust Principal Component Analysis (RPCA) to separate the background information and defect information of the solar cell defect picture and judged the defect through the decomposed defect information. This method requires a template. If the production batch is inconsistent and the illumination of the light source is uneven, the template needs to be re-selected. Therefore, the adaptability is limited. The above-mentioned traditional feature extraction methods rely heavily on the selection of artificial features, resulting in limited adaptability of the defect detection of solar cells under complex backgrounds.

Solar cell surface defects under visible spectrum are various, including broken gates, paste spot, thick lines, dirty cell, missing corners, scratches, color differences, etc. The non-uniform background, complex textures and Non-Gaussian color distribution weaken defect discernable feature. The gaps of color and lattice between different polysilicon cells are also quite large. This leads to the existing manually image feature extracting methods are difficult to effectively perceive the multiple types of defect information in different light spectrum range. The reason lies in the fact that the defect features of the algorithms depend on manual selection, which is hard to represent some features in the multispectral images.

Deep learning uses a large amount of data to train deep learning models, including a large number of low and high-level features. Wang et al.[24] applied deep confidence neural networks to the detection of cracks and missing corners of solar cells. The deep belief network is an unsupervised learning method that can reconstruct a defect-free model based on the current image. However, the number of data sets used in this method is small. Moreover, there have been no reports about surface defect detection of solar cells using deep learning.

In recent years, the CNN and its variants have been preliminary studied in the field of surface defects detection such as textiles, strip steel, and buildings [25]. Weimer [26] et al. used the deep convolutional neural networks to detect surface defect datasets such as textile and steel in 2016. This paper discussed the effect of depth and width of the CNN model on test results. Wang et al. [27] proposed a new deep CNN model structure in 2017. The model uses all types of defect-free and defect samples together as input, and the output is a 12-class classifier: 6 non-defective and 6 defectives. However, the dataset is small and may have problems of overfitting. In order to solve the problem that there is not enough labelled data in the defect detection, Kim [28] and others proposed a defect detection algorithm based on transfer learning. The paper transferred the weight parameters of other models to the current defect detection model to achieve sharing of weights and easing overfitting. Lin et al. [29] first apply a convolutional neural network to the LED surface and realized the identification and positioning of various defects. And the accuracy reaches 94%. However, the datasets which are studied in the above several literatures are single-channel images and it is difficult to describe and deal with the multi-spectral characteristics of complex surface defects in solar cells. As to solar cell test, there are a few researches based on CNN. Pierdicca, R., et al. [30]. used convolutional neural networks to detect remote sensing images of solar cells and identify broken cells in the solar cell module. The author successfully applies CNN to solar cell defect detection. The disadvantage is that the precision of CNN in this paper is about 70% due to the low-resolution remote sensing images of solar modules. S Deitsch et al. [31] applied a convolutional neural network for EL image detection of solar cells and was able to detect various EL defects. Comparing with the traditional machine vision method, the algorithm in this paper achieves 88.36% accuracy on the dataset, which increase by 6 percentage points. At the same time, the detection speed of the algorithm meets the requirements of real-time production.

To achieve the defect inspection of solar cell surface, we have to deal with two major problems. One problem is to significantly highlight multiple defects characteristics by employing multiple spectrum information. The other is automatic multi-spectrum feature extraction and inspection of solar cell surface.

In this paper, focusing on the visual intelligent detection of surface defects in polycrystalline silicon cells based on deep learning, the high dynamic camera is used to collect the multi-spectral images of solar cells. Then the defect dataset of solar cells is established. Next, the optimized design of solar cell convolutional neural network model is achieved. Finally, a multi-spectral convolutional neural network model is proposed based on the CNN optimization model. The detection accuracy and feature extracting ability are significantly improved. The paper has the following research contributions.

1. The effects of model depth and convolution kernel size variation are evaluated and analyzed in this paper. The solar cell CNN model with optimized CNN model depth and convolution kernel size are established, which can better distinguish multi-defect features.

2. Based on the selected solar cells CNN model, a multi-spectral solar cell CNN network model is proposed so as to extract the multi-spectrum features of solar cell surface. The comparison of Multispectral solar cell CNN models and solar cell CNN models is experimentally analyzed. The cross-validation results prove that the model is robust and adapts to various types of random feature defects and has strong ability to resist over-fitting.

3. Experiments with multiple classifications are carried out. Experimental results show that error detection occurs more between positive and negative samples. Furthely, the features of the middle layer are displayed to enhance the interpretability of the model.

To the best of our knowledge, this is the first paper which solves the solar cell surface defect inspection using a deep learning approach.

The rest of the paper is organized as follows. In Section 3, the defect dataset of solar cells is established then visual acquisition system of multi-exposure welding images is designed. Furthermore, a multi-spectral solar cell CNN network model is proposed. In Section 4, several comparative experiments including traditional machine learning algorithms, solar cells CNN and Multispectral solar cell CNN are performed, demonstrating the adaptability and robustness of Multispectral solar cell CNNs to complex non-uniform surfaces. Finally, concluding remarks are given in Section 5.

## 3  Methods

In this section, the multi-spectral characteristics of solar cell surface defects are analyzed, and defect datasets are established. Then the solar cell CNN model and the multi-spectral solar cell CNN model are designed. The effect of model depth and convolution kernel size variation on the detection performance is discussed. The solar cell CNN model structure with the best performance is selected. Furthermore, a multi-spectral solar cell CNN model is proposed to improve the proposed solar cell CNN.

### 3.1. Defect feature analysis and dataset

#### 3.1.1 Multispectral defect feature analysis

Solar cells appear a complex texture background including irregular lattice features, and grid line features. The shape and location of lattice are random, whose color is similar to background color of solar cell. The grid line is the energized current-conducting part of the cell, which is silver white. The surface defects of solar cells in the visible light spectrum range include chipping, broken gates, leaky paste, dirty sheets, scratches, thick lines, and chromatic aberrations. The shape, size and spectrum characteristics of each defect show a big difference. Some typical defects are shown in Fig. 1. Broken gate refers to the breakage and loss of the printed finger lines on the surface of the cell. Paste spot is the dripping of the paste when the cell sheets are printed the grid. Dirty cell refers to large dust or dirt on the solar cell. The thick line indicates that the printed weight of the cell sheet is too heavy and the thickness of the gate line is uneven. Scratches are caused by a sharp object passing over the cell. Destruction refers to the collapse of the blue coating on the edge of the cell, which is generally white. Chromatic aberration is due to firing problems with rainbow colors.

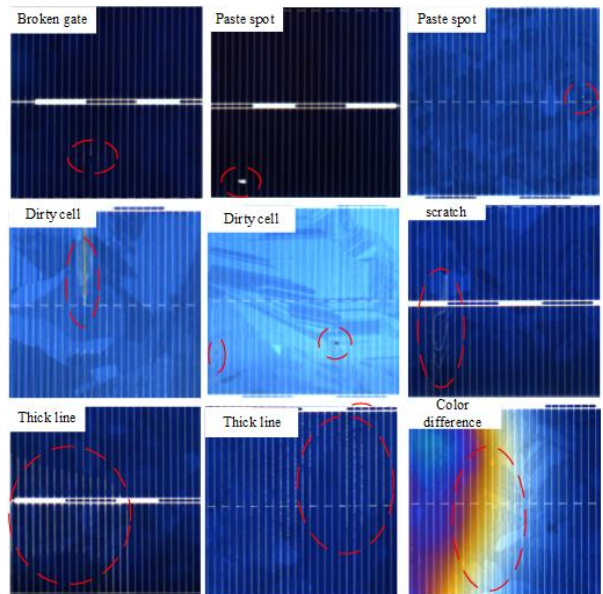

Fig.1 Various surface defects of solar cell

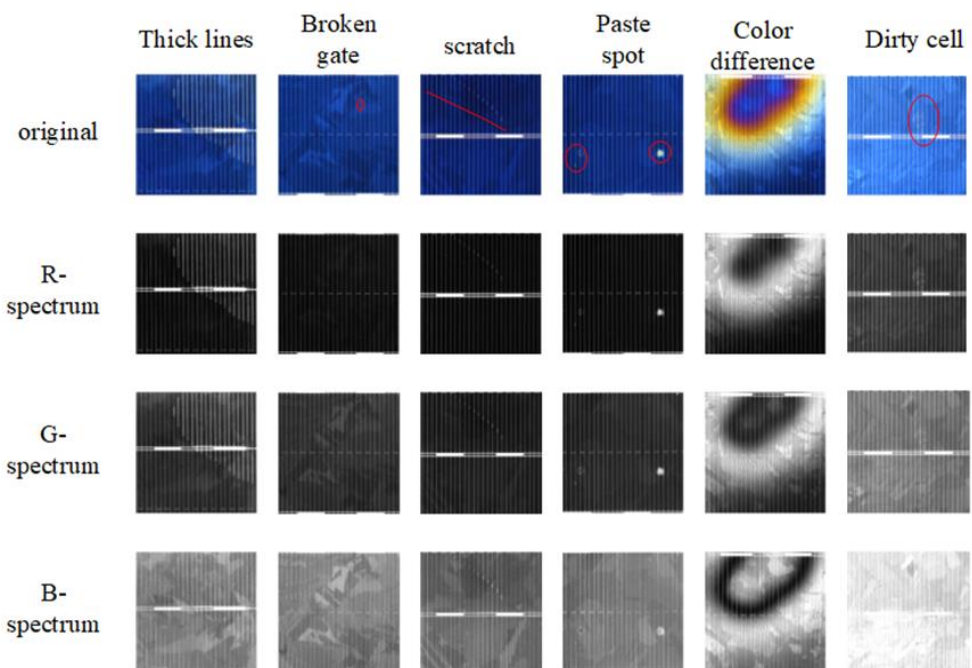

Fig.2 Defection in different spectral

The characteristics of solar cell surface defects in different spectrum are shown in Fig. 2. It can be seen that different defects have different contrast in the same spectrum, and the same kind of defect also show different contrast in different spectra. For example, the broken gate, thick line, and dirty cell are more obvious in the red or green spectrum image, while the blue spectrum has poor contrast due to the interference of the lattice background. The paste spot, and color difference are all evident in the three spectra; The scratches are more pronounced in the green and blue spectrum.

*3.1.2 Multispectral defect dataset*

The actual size of the solar cell used in this paper is 156mm×156mm. The images are collected by a 5 million color camera. After extracting the region of a solar cell, color images with a height and width of 1868×1868 pixels are obtained. During training process, the image needs to be chunked in order to ensure the speed of model training and retain defect information.

Using slide-splitting [26] [27] to segment the images into smaller pictures can both expand the dataset and highlight the defect information. Sliding segmentation is widely used in the establishment of deep learning datasets and has a good effect for extending scale, reducing overfitting of data training. The sliding window has a size of 469×469 pixel and moves along the rows and columns over the whole image with a 235-pixel stride. The steps of the sliding-splitting division are as shown in Fig. 3, it takes image blocks of size 469×469 pixel extracted from the original 1828×1828-pixel images as the input. In this way, we can extract 49 small blocks from one original image. After splitting each segmented cell is manually screened and classified. Finally, 15330 undefective images and 5915 defective images are obtained. The types of defects include broken gates, paste spot, dirty cell, thick lines, scratches, and color differences. The specific number of each defect is given in Table 1.

*3.2. Solar cell CNN model structure*

Based on the dataset, the three challenging problems about training and inspection of solar cell surface defect mainly include：1) There are 6 types of defects in the dataset. The characteristics of each defect type are quite different in shape, scale and spectrum; 2) the lattice shape of the polycrystalline silicon cell is random. The background texture features of cells are relatively complex, and the defect features cannot easily be discerned; 3) the surface color of the solar cell is a random non-Gaussian distribution, which causes the random and non-uniform brightness of the solar. Next, solar cell convolutional neural network model will be designed.

Table 1 The type and number of defection dataset

| defect | Broken gate | Paste spot | Dirty cells | Thick lines | scratches | Color difference |
|---|---|---|---|---|---|---|
| amount | 1330 | 1790 | 1830 | 361 | 350 | 254 |

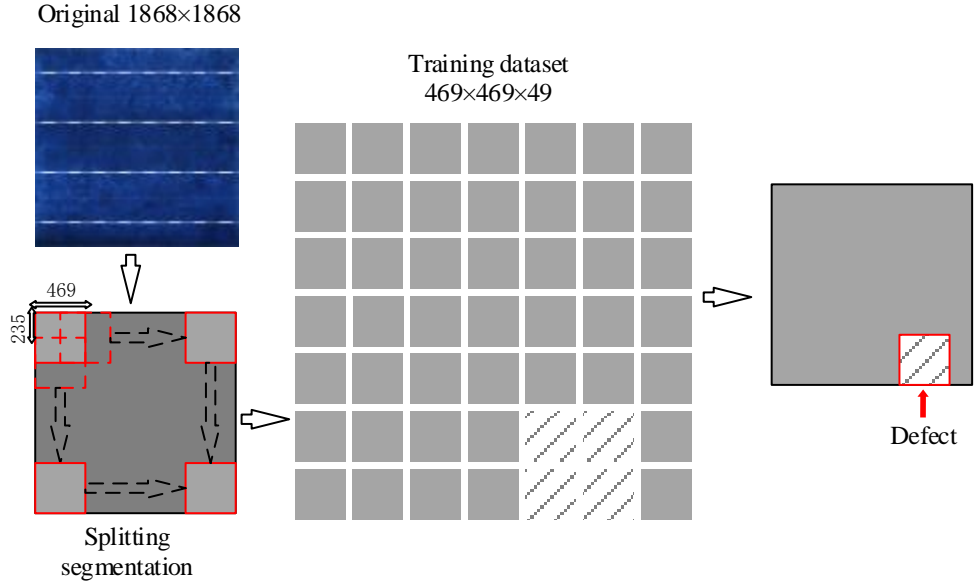

Fig. 3 Creating dataset by slid-splitting

*3.2.1. Solar cell CNN model design*

Convolutional neural network (CNN) is a learnable end-to-end model that combines feature extraction and inspection. It captures the low and high-level features of images through multiple layers of convolution and pooling, then classifies through the full connection layer. A complete CNN model includes Input, Conv, Activation, Pool, and Fully Connection.

The convolutional layer (Conv) convolutes on the input image by moving the convolution kernel to extract image features. The convolution kernel is a square filter. The filter size is optional, usually 3x3, 5x5 or 7x7.

Pooling is also called down-sampling. The effect is to reduce the size of the feature map output by the intermediate process without changing the dimension of the image, so as to reduce the image size and reduce the complexity of the model training calculation. Generally, the calculating methods of the pooling layer include the average pooling, the max pooling, and the Gaussian pooling, among which the max pooling is more suitable for the extraction of image texture detail features.

Fully Connection: As the last part of the convolutional neural network model, Fully Connection is connected to the Softmax classifier. Each output of the fully connected layer can be considered as multiplying each node of the previous layer by a weight coefficient and finally adding a bias value. The last layer of the fully connection produces the output of entire network. The output has the same number of K neurons as the input label. $z_i$ represents the input of the Softmax classifier while $p(z_i)$ represents the output probability.

$$p(z_i) = \frac{exp(z_i)}{\sum_{i=1}^{K} exp(z_i)} \tag{1}$$

The above layers are stacked together to form a complete CNN model. The optimal objective function is as follows

$$\theta^* = argmin \frac{1}{T} \sum_t L(\int_\theta (x^{(t)}, y^{(t)}) + \lambda \Omega(\theta)) \tag{2}$$

Where, $\theta^*$ are the set of final parameters while λ represents learning rate. $x^{(t)}$ and $y^{(t)}$ indicate the input data and corresponding labels of the network model. Input data $(x^{(t)}, y^{(t)})$ into the network for supervised training and decision making. $\theta = \{w_1, b_1, ..., w_{m+1}, b_{m+1}\}$ represents weights (network weights $w$ and $b$) of all layers in the network model. $\Omega(\theta)$ represents regularization hyperparameter used to penalize excessively high network weights to prevent overfitting. CNN uses the difference between the output and the tag to control the change of weight and uses the stochastic gradient descent (SGD) to solve the optimal value of the back propagation of the network model.

Activation: Since a linearly structured network cannot fit complex functions, the activation function layer must be a non-linear function. Typical activation functions are sigmoid, tanh, ReLU and their variants. Among them, ReLU is selected as the activation function in the paper due to its best performance in the ImageNet recognition contest [32]. It can speed up the random gradient dropping, and its function expression is as follows:

$$y = \begin{cases} x, x \geq 0 \\ 0, x < 0 \end{cases} \tag{3}$$

In order to reduce computation time and memory consumption when training the model, solar cell images in dataset are resized to 256×256×3. During model training, $L_2$ regularization and Dropout [33] are used to prevent overfitting of model training. Dropout refers to the inactivation of some hidden layer nodes of the network at the time of model training. These deactivated nodes do not work in the current training. During convolution and pooling operations, extra padding is performed in the image boundary area to ensure the integrity of images. At the same time, because the features of the defects

**5 / 14**

are mostly edge-based, max-pooling is chosen as the pooling method of the model. Compared with average pooling, the maximum pooling can better preserve the texture features.

The solar cell CNN models can extract image features autonomously, but solar cell surface defects have a big difference in different spectra. Therefore, a new model structure is designed based on the Alexnet model. Adjusting the convolution kernel size and network depth to enhance the model's defect discrimination capabilities. The solar CNN model structure is shown in Fig. 4. The different convolution kernel size and feature output sizes of the two models are shown in Table 2.

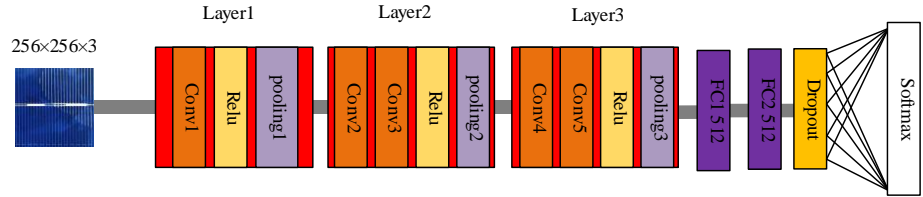

Fig.4 Structure of solar cell CNN

Table 2 Different structure of solar cell CNN

|  | 1th structure | 2th structure | 3th structure |
| --- | --- | --- | --- |
| Layer1 | Conv1-3×3 | Conv1-3×3 | Conv1-7×7 |
| Layer2 | Conv2-3×3 | Conv2-3×3 | Conv2-5×5 |
|  |  | Conv3-3×3 | Conv3-5×5 |
| Layer3 | Conv3-3×3 | Conv4-3×3 | Conv4-3×3 |
|  |  | Conv5-3×3 | Conv5-3×3 |

*3.2.2 The depth and convolution kernel size selection of solar cell CNN*

The depth and convolution kernel size of the CNN model have a significant influence on the test results [26]. The depth of the model refers to the number of convolution layers in the model. Changes in the kernel size in the convolutional layers also affect the characteristics extracted from the image. The deeper CNN can be, the more advanced the features extracted; the larger the convolutional kernel is, the more surrounding information the extracted features contain. Referring to the literature on the use of CNN models to deal with surface defects in recent years [27][28]**Error! Reference source not found.**, the influences of depth and width changes of CNN models on the final results are significant. According to the above literature, the following three model depths and convolution kernel sizes are compared. Finally, the best-performing model structure is determined for subsequent multi-spectral solar cell CNN model experiments. The three model structures are shown in Table 2, in which Conv$i$-$j\times j$ ($i$=1,2,3,4,5. $j$=3,5,7) represents the $i$-th convolution layer, and the convolution kernel size is $j\times j$.

*3.3. The Multispectral solar cell CNN model structure*

Some defects are prominent in some spectra while weak in others. Solar CNN models can extract features in the mixed spectrum. Some defect features of solar cells are located on the gate lines, such as thick lines, broken gates, etc., while the others are in the background, such as scratches, dirty cells, paste spots, etc.

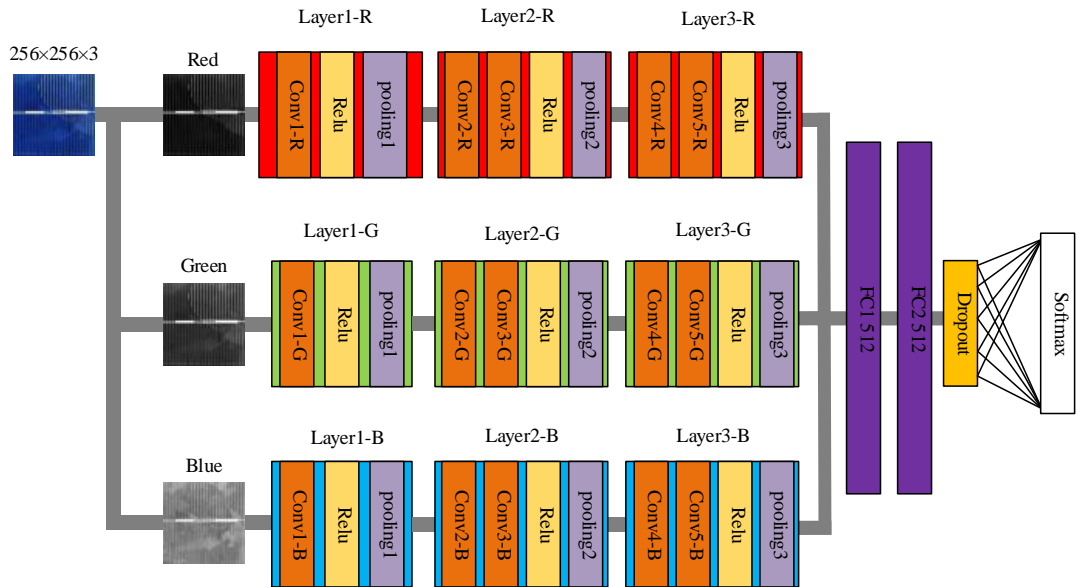

Fig. 5 Multispectral solar cell CNN structures

Table 3 The architecture and parameters of MS-CNN

| Name | kernel | Solar cells CNN | | Multispectral solar cell CNN | | | |
|---|---|---|---|---|---|---|---|
| | | structures | output | Red-spectral | Green-spectral | Blue-spectral | Output |
| Layer1 | 16@7×7 | Conv1 | 256×256×16 | Conv1-R | Conv1-G | Conv1-B | 3@256×256×16 |
| | 2×2 | Pool1 | 128×128×16 | Pooling1-R | Pooling1-G | Pooling1-B | 3@128×128×16 |
| | 32@5×5 | Conv2 | 128×128×32 | Conv2-R | Conv2-G | Conv2-B | 3@128×128×32 |
| Layer2 | 32@5×5 | Conv3 | 128×128×32 | Conv3-R | Conv3-G | Conv3-B | 3@128×128×32 |
| | 2×2 | Pool2 | 64×64×32 | Pooling2-R | Pooling2-G | Pooling2-B | 3@64×64×32 |
| | 64@3×3 | Conv4 | 64×64×64 | Conv4-R | Conv4-G | Conv4-B | 3@64×64×64 |
| Layer3 | 64@3×3 | Conv5 | 64×64×64 | Conv5-R | Conv5-G | Conv5-B | 3@64×64×64 |
| | 22 | Pool3 | 32×32×64 | Pooling3-R | Pooling3-G | Pooling3-B | 32×32×192 |
| FC1 | 512 | FC1 | | FC1 | | | |
| FC2 | 512 | FC2 | | FC2 | | | |
| Softmax | 2 | Softmax | | Softmax | | | |

Therefore, aiming at the characteristics of distinguishing degree of solar cells surface defects in different spectra, a multi spectral solar cell CNN model is established by extracting three basic networks. The three spectra in the original image are separated and sent to different convolutional neural networks. The output characteristics of the three networks are then connected and fed into the fully connected layer, ultimately producing a predictive output. The multi-spectral solar cell CNN model structure is shown in Fig. 5 and Table 3. There are three parallel feature extraction layers. Conv-R, Conv-G and Conv-B represent convolutional layers of different spectra. ReLU is a nonlinear activation function; Pooling is the maximum pooling, that is, the features of multiple spectra of images. The Multi-spectral solar cell CNN is composed of maximum value of the feature points in the neighborhood is selected as the output. In the multispectral solar CNN model, mixed features of multiple spectra are extracted. The output dimension of the third convolutional layer is 32x32x64, which contains all the feature information extracted from multiple spectra and is then input to the fully connected layer. The multi-spectral solar cell CNN separates the three spectra of the color image into the convolution pool after inputting the image, and outputs the feature results Layer3_B, Layer3_R, and Layer3_G in the third convolutional layer of the model and combines the three feature results to obtain Layer3. The output feature size is 64×64×192. The output features' size is 64 x 64 x 192, which are then input into the fully connected layer and the final Softmax layer for inspection. Extracting the feature information of the images from the three spectra, since the features of each defect in the different spectra are very distinct, the finally obtained image features are more distinguishable and conducive for defect detection.

## 4  Experimental results analysis and discussion

The experiment is completed on the Ubuntu 16.04 platform using the TensorFlow framework. The computer's CPU used for training is the Core i7 series, with 32GB memory and two GTX1080 graphics cards. The learning rate of the CNN model and the multi-spectral solar cell CNN model is both chosen as $\lambda=0.0001$, and the epochs of training is 10,000. The Dropout neuron ratio is 50%. The experiments in this article are mainly divided into the following three parts.

*4.1. Selection of CNN depth, kernel size and stride step*

The design of the CNN model is closely related to different defect datasets. Though the CNN models that detect different defects are similar, the model structures including the depth and the size of the convolution kernel vary. This paper designs three CNN models with different convolutional depths and kernel sizes and selects the best CNN model for solar cells. The training images are fed into the network. Then the trained network is loaded for the test. Each type of defect and non-defective images in the dataset are randomly selected as a test set by 20%, and the test set for the three experiments is the same.

To more accurately evaluate the vague scratches detection results, we use precision, recall rate and F-measure. Precision measures the exactness or fidelity of detection and segmentation and is calculated in Eq. (4). Recall describes the completeness of detection and segmentation and is defined in Eq. (5). F-measure combines precision and recall and is computed in Eq. (6). Table 4 shows the precision, recall, and F-measure for the solar cell CNN. (TP represents a true positive, that is, images labeled as defective are correctly detected; FP indicates false positives, that is, images labeled as good are erroneously detected as defective; FN means false negative, that is, images labeled as defective are erroneously detected as non-defective; TN represents a true negative, that is, images labeled as non-defect are correctly detected as non-defect)

$$Precision = \frac{TP}{TP + FP} \tag{4}$$

$$recall = \frac{TP}{TP + FN} \tag{5}$$

$$F - measure = \frac{2 \times precision \times recall}{precision + recall} \tag{6}$$

The experimental results for the three different structures of solar CNN are given in Table 4, and each experimental result is the average of the 5-fold cross-validation. The experimental results show that the precision of the solar cell CNN model with 5-layer convolutional layer is 2% higher than that of a 3-layer convolutional layer CNN model. When the kernel size is increased on the basis of the five-layer convolutional layer CNN model, both precision and recall are improved, but they are all within 0.6%, and the effect is not significantly improved. If the depth of the convolution layer and kernel size are increased the training time of the model will be prolonged and the training pressure will be increased. Therefore, the third model structure is selected as the solar cell CNN model and subsequent experiments are conducted.

It is considered that deeper network has better ability to extract high-level features and increasing the width of the network can also improve the performance of the network. At the same time, the larger convolution kernel in the low-level feature map has a larger receptive domain, which has a better effect on extracting large-area feature defects [26].

Table 4 Results of different structures of Solar Cell CNN

|  | 1th structure | 2th structure | 3th structure |
| --- | --- | --- | --- |
| precision（%） | 85.11 | 86.16 | 87.30 |
| recall（%） | 96.00 | 96.46 | 97.04 |
| F-measure | 0.9022 | 0.9041 | 0.9187 |

After the network structure is selected, an experiment to determine the sliding window step size is performed. The size of the image in the dataset is 469x469, which contains exactly the main grid lines on either side, or one main grid line in the center. Therefore, to ensure the stability of all images in the dataset, we choose the quarter length of the original image as the size of the dataset image. In order to ensure the validity of dataset, three sets of images of different sizes are prepared and tested separately. Table 5 shows the results of three different strides based on five-fold cross-validation. The experiment is based on solar CNN.

Table 5 Results of different strides of Solar Cell CNN

|  | 234×234 | 469×469 | 623×623 |
| --- | --- | --- | --- |
| precision（%） | 76.86 | 87.30 | 76.85 |
| recall（%） | 95.88 | 97.04 | 81.02 |
| F-measure | 0.8532 | 0.9187 | 0.7888 |

As can be seen from Table 5, if the splitting stride is too long, many features will be lost, so the classification effect is relatively poor. As for the smaller image dataset, the detection result is also poorer than the middle one. Furtherly, a serious overfitting occurred on the test dataset. The reason is that defective solar cells account for only about 2% of the total production. If the original images are split too small, it will not only affect the recognition speed of the whole battery, but also lead to further imbalance of the proportion of defective samples, resulting in more serious over-fitting. Therefore, Therefore, too small images' segment are also not conducive to the detection of defects.

*4.2 Comparison and Analysis between multispectral solar cell CNN model and solar cell CNN*

The Multispectral solar cell CNN is based on the solar cell convolutional neural network model and analyzes the characteristics of different solar cell surface features defects under different spectra and improved the obtained network structure. To increase the credibility of the training results, this experiment firstly used K-fold cross-validation to traverse all the images to verify that the defect detection result of solar cell CNN has higher reliability [33]. The K-fold cross-validation experimental procedure is as follows: The data set is divided into K sub-samples, a single sub-sample is reserved as a test set, and another (K-1) sample is used as a training set. The experiment is performed K times, and the average is taken as the final test result. This paper used a 5-fold cross validation.

By analyzing the performance characteristics of various defect types of the cell under different spectra, it is found that defects such as broken gate, paste spot, thick line, dirty cell, color difference, scratch, have different performance characteristics in different spectra. Some defects are more pronounced in a particular spectrum while they are reduced in other spectra. The multi-spectral solar cell CNN model separates the three spectra of the image to make the features extracted by the model more distinguishable and reduce the false detection rate of defects. In order to verify the effectiveness of the multi-spectral feature separating's extraction, we compared the experimental results of the solar cell CNN model and the multi-spectral solar cell CNN model and used a 5-fold cross validation to traverse all the images. Table 6 shows the 5-floder training and testing dataset.

Table 6 The division of each experiment

| type | training set | test set | sum |
| --- | --- | --- | --- |
| good | 12264 | 3066 | 15330 |
| broken gate | 1064 | 266 | 1330 |
| paste spot | 1432 | 358 | 1790 |
| dirty cell | 464 | 366 | 1830 |
| thick line | 289 | 72 | 361 |
| scratches | 280 | 70 | 350 |
| color difference | 203 | 51 | 254 |

Then the statistics of the detection performance of each type of defect is used to evaluate the performance of various defects by using solar cell CNN and Multi-spectral solar cell CNN. Each time a certain type of defect in the dataset is used to test separately. In order to fully demonstrate the experimental process, Table 7 shows three indicators for each fold of each defect in the five-fold cross-validation, and the average of all the indicators after five experiments.

In Table 7, it can be seen that the experiment results show that the multi-spectral solar cell CNN model has improved the detection rate of dirty cell, thick line, broken gate, color difference, and paste spot in the detection of cell defects. Among them, the dirty cell, and broken gate is increased by more than 1%, the scratch detection rate decreased by about 0.5%. The correct detection rate of the non-defective cells is increased by 1.3%. From the longitudinal comparison of five experiments, the correct detection rates of color difference, and dirty cell in the five experimental results of the multi-spectral solar cell CNN model are higher than that of the CNN model. The correct detection rate of five training non-defective images has increased by about 1% to 4%. The defects of thick lines, broken gates, and paste spot are higher than the CNN model's detection results three times in the five experimental results. At the same time the average detection rates are higher. It should be noted that for defect detection, high recall means fewer defects missing. As it can be seen from Table 7, the recall of multi-spectral CNN is relatively higher. In addition, from the comparison of the data of the 5-fold cross-validation, it can be seen that there is no large fluctuation in results of each experiment. This also confirms that the multi-spectral CNN model avoids over-fitting well and has better generalization ability. The detection rate of scratches is also higher than the solar cell CNN model three times in the experiment, but the average detection rate is reduced by 0.5%, and the detection rate is lower than other defects. Based on the final test results, the multi-spectral convolutional neural network is better for detecting defects in solar cells.

All experimental results in Table 6 are counted and the average number of all samples was calculated. The confusion matrix for the multispectral CNN experimental average results is given in Table 8.

Table 8 Test sets' result of multi-spectral solar cell CNN

|  | Non-defect | thick line |
|---|---|---|
| Non-defect | 3017 | 49 |
| thick line defect | 10 | 62 |
|  | Non-defect | broken gate |
| Non-defect | 3015 | 51 |
| broken gate | 46 | 220 |
|  | Non-defect | scratches |
| Non-defect | 3006 | 60 |
| scratches | 20 | 50 |
|  | Non-defect | paste spot |
| Non-defect | 3032 | 34 |
| paste spot | 43 | 315 |
|  | Non-defect | color difference |
| Non-defect | 3023 | 43 |
| color difference | 1 | 50 |
|  | Non-defect | dirty cells |
| Non-defect | 2998 | 68 |
| dirty cells | 18 | 348 |

The convolutional neural network can autonomously extract low and high-level features of the image itself. The convolution output after the first layer is the low-level features of the image, and the output after the convolution of the last layer is the high-level feature. Fig.6 shows the output of the low-level and high-level features of the solar cell CNN model and the multi-spectral solar cell CNN model. Fig.6 (a) is the defect image, Fig.6 (b) and Fig.6 (d) are the low-level features of the output of Layer1 of the two models, and Fig.6 (c) and Fig.6 (e) are the high-level features of the output of Layer3. It can be seen in Fig. 2 that the characteristics of the dirty cell in the red and green spectrum are significant, but the performance in the blue spectrum is not obvious. As can be seen in Fig.6(d), the multi-spectral solar cell CNN model has obvious defect features in low-level features extracted from the red and green spectra. Most of the features extracted in the blue spectrum are lattice features, the defect features are not obvious, and most of the output feature images are pure black. Fig.6 (e) shows the high-level features output of Multispectral solar cell CNN. The defect features extracted from red and green spectra are more accurate and the gray values are higher. The features of lattice and grating lines in the blue spectrum are obvious and there are few defect features.

Comprehending low and high-level features extracted by multi-spectral solar cell CNN, it can be found that the background texture feature and defect feature of the image are well separated. For dirty cell defects, the features in the red and green spectrums are more pronounced. The multi-spectral solar cell CNN separates the defect features and background features of the dirty cell. Compared with the hybrid spectral feature extraction of solar cell CNN model, the feature extraction ability of multi-spectral solar cell CNN model is strengthened, and the extracted feature is more distinguishable.

Thus, multi-spectral solar cell CNN model has better adaptability to the different characteristics of solar cell surface defects. And as to other defection like thick line, paste spot or scratches, their features are also more illustrate the adaptability of unknown multi-spectral solar cells samples in some extent.

Next, the ratio of the training and test sets is 8:2, 6:4, and 4:6 respectively, as is shown in Table 8. This experiment is conducted to demonstrate that multispectral solar cell convolutional neural networks are still effective when the data set is still a small percentage of overall production data. Table 9 shows the results of three experiments.

Table 9 Results for three different scale datasets

|  | 8:2 | 6:4 | 4:6 |
|---|---|---|---|
| precision（%） | 88.41 | 87.53 | 83.41 |
| recall（%） | 98.40 | 96.80 | 96.06 |
| F-measure | 0.9401 | 0.9193 | 0.8929 |

From Table 9 it can be obtained that as the ratio of the training and test sets increases, the precision, recall and F-measure of multi-spectral solar cell CNN increases slightly. When the ratio of test set to verification set is 4:6, precision is reduced by about five percentage points. The experimental results illustrate the adaptability of unknown multi-spectral solar cells samples in some extent.

Table 7 The results of CNN experiment

| | | | Thick line | Broken gate | Scratches | Paste spot | Color difference | Dirty cells |
|---|---|---|---|---|---|---|---|---|
| 1 | Solar CNN | precision | 80.57 | 82.32 | 78.56 | 91.30 | 92.00 | 96.97 |
| | | recall | 93.83 | 93.95 | 93.68 | 94.51 | 94.55 | 94.47 |
| | | F-measure | 0.87 | 0.88 | 0.85 | 0.93 | 0.93 | 0.96 |
| | Multispectral solar cell CNN | precision | 77.78 | 83.00 | 72.85 | 87.90 | 96.10 | 99.95 |
| | | recall | 97.55 | 97.70 | 97.39 | 97.83 | 98.01 | 98.09 |
| | | F-measure | 0.87 | 0.90 | 0.83 | 0.93 | 0.97 | 0.99 |
| 2 | Solar CNN | precision | 86.08 | 85.64 | 71.38 | 79.80 | 96.10 | 91.99 |
| | | recall | 98.10 | 98.09 | 97.71 | 97.95 | 98.29 | 98.22 |
| | | F-measure | 0.92 | 0.91 | 0.82 | 0.88 | 0.97 | 0.95 |
| | Multispectral solar cell CNN | precision | 88.87 | 86.47 | 82.86 | 87.98 | 96.10 | 95.60 |
| | | recall | 98.56 | 98.52 | 98.46 | 98.54 | 98.67 | 98.66 |
| | | F-measure | 0.93 | 0.92 | 0.90 | 0.93 | 0.97 | 0.97 |
| 3 | Solar CNN | precision | 83.30 | 78.22 | 57.12 | 84.00 | 100.00 | 91.70 |
| | | recall | 97.43 | 97.26 | 96.29 | 97.45 | 97.85 | 97.66 |
| | | F-measure | 0.90 | 0.87 | 0.72 | 0.90 | 0.99 | 0.95 |
| | Multispectral solar cell CNN | precision | 90.23 | 79.68 | 61.42 | 85.15 | 100.00 | 93.34 |
| | | recall | 98.77 | 98.61 | 98.21 | 98.70 | 98.89 | 98.81 |
| | | F-measure | 0.94 | 0.88 | 0.76 | 0.91 | 0.99 | 0.96 |
| 4 | Solar CNN | precision | 93.01 | 79.69 | 68.55 | 85.15 | 100.00 | 91.50 |
| | | recall | 97.09 | 96.62 | 96.09 | 96.83 | 97.29 | 97.04 |
| | | F-measure | 0.95 | 0.87 | 0.80 | 0.91 | 0.99 | 0.94 |
| | Multispectral solar cell CNN | precision | 90.23 | 78.95 | 70.00 | 88.47 | 100.00 | 92.87 |
| | | recall | 98.48 | 98.27 | 98.05 | 98.45 | 98.63 | 98.53 |
| | | F-measure | 0.94 | 0.88 | 0.82 | 0.93 | 0.99 | 0.96 |
| 5 | Solar CNN | precision | 86.08 | 86.47 | 85.6 | 91.5 | 96 | 95.8 |
| | | recall | 96.86 | 96.87 | 96.84 | 97.04 | 97.18 | 97.17 |
| | | F-measure | 0.91 | 0.91 | 0.91 | 0.94 | 0.97 | 0.96 |
| | Multispectral solar cell CNN | precision | 86.13 | 84.57 | 71.38 | 89.84 | 96 | 94.24 |
| | | recall | 97.48 | 97.43 | 96.97 | 97.58 | 97.73 | 97.69 |
| | | F-measure | 0.91 | 0.91 | 0.82 | 0.94 | 0.97 | 0.96 |
| AVR | Solar CNN | precision | 85.81 | 82.47 | 72.24 | 86.35 | 96.80 | 93.59 |
| | | recall | 96.68 | 96.55 | 96.08 | 96.70 | 97.04 | 96.94 |
| | | F-measure | 0.91 | 0.89 | 0.82 | 0.91 | 0.97 | 0.95 |
| | Multispectral solar cell CNN | precision | 86.64 | 82.53 | 71.70 | 87.87 | 97.62 | 95.20 |
| | | recall | 98.19 | 98.10 | 97.82 | 98.21 | 98.39 | 98.35 |
| | | F-measure | 0.92 | 0.90 | 0.83 | 0.93 | 0.98 | 0.97 |

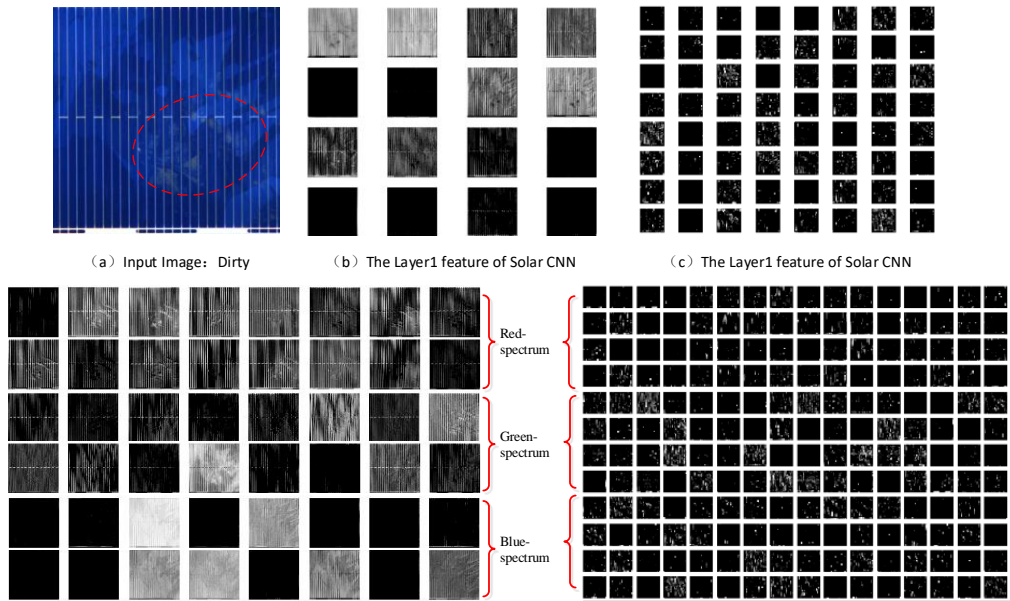

Fig. 6 The feature of solar cell CNN and multi-spectral solar cell CNN

### 4.3 Multi-Class Comparison and Analysis of Multispectral solar cell CNN Models and Some Typical Machine Learning Methods

To furtherly validate the stability of multispectral CNN for different defects, a multi-class experiment is performed the experimental results of solar CNN and multispectral solar cell CNN are shown in Table 10. The data in Table 10 is the average of the five-fold cross-validation.

As can be seen from Table 10, in the multi-classification task, the accuracy of the multi-spectral CNN is 2 to 6 percentage higher than solar cell CNN for each defect. At the same time, misidentification mainly occurs between positive and negative samples compared to defect classes. At the same time, misidentifications mainly occur between positive and negative samples compared to the types of defects. Moreover, compared with the accuracy rate of about 85 percent of the positive and negative samples of the two classifications, the precision of the multi-class is on average 8 percent lower. The accuracy of multi-class classification of defects is lower than binary classifications maybe because the dataset is not balanced. For the specific defect types like scratches or color difference, too few images of certain defective solar cells are an important cause of the decline in multi-classification effects. But in industrial manufacturing, different types of defective solar cells will be collected and reduced to the defective grade. Therefore, binary classification can improve the efficiency of manufacturing.

### 4.4 Comparison and Analysis of Multispectral solar cell CNN Models and Some Typical Machine Learning Methods

The surface defects of solar cells are various, the background is complex, and the features of manual extraction are more difficult. Before the large-scale application of convolutional neural networks, the method of manually extracting features for defect detection is widely used in machine vision. LBP+HOG-SVM and Gabor-SVM have good results in the field of surface defect detection[16][35], Among them, LBP stands for Local Binary Pattern, HOG stands for Histogram of Oriented Gradient, and SVM stands for Support Vector Machine. Like the Gabor feature, they are feature descriptors used in traditional machine vision methods. Therefore, the above two commonly used machine learning methods are selected as comparative experiments. Table 11 includes the recall and precision for the four methods of LBP+HOG-SVM, Gabor-SVM and traditional CNN models, Multispectral solar cell CNN models.

The parameters of LBP+HOG-SVM are as follows: the original images are splitting into $12 \times 12$ regions to calculate LBP features with a radius of 1 in 8 neighborhoods. As to Gabor, the cell size is $8 \times 8$ while the block size is $2 \times 2$ and the overlap is $1 \times 1$. The parameters of Gabor-SVM are as follows, the down-sample image' size is $10 \times 10$, the Gabor kernel size is $31 \times 31$, Gabor kernel's energy preserving ratio is 0.9. Gabor kernel's number is 40, which is in 5 scales and 8 orientations.

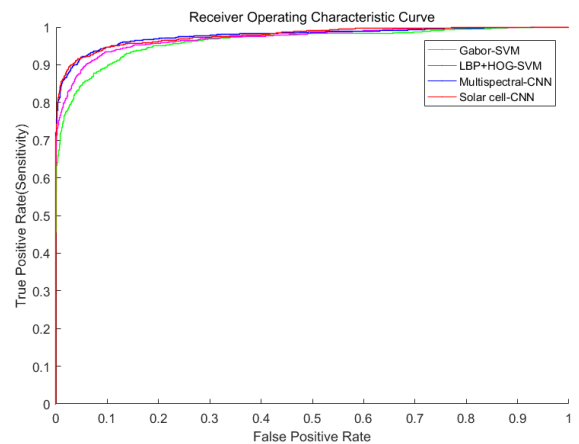

Fig. 7 The Receiver Operation Characteristic (ROC) curves

The experimental results are gained by using a 5-fold cross validation experiment. Fig.7 shows the receiver operation characteristic curves for the four detection methods.

Table 10 Multi-class experimental results of solar cell CNN and multi-spectral CNN

|  |  | Thick line | Broken gate | Scratches | Paste spot | Color difference | Dirty cells | Right |
|---|---|---|---|---|---|---|---|---|
| Solar cell CNN | Thick line | **73.61** | 0 | 0 | 0 | 0 | 0 | 26.39 |
|  | Broken gate | 0.38 | **78.20** | 0 | 0.38 | 0 | 0 | 21.43 |
|  | scratches | 0 | 1.42 | **40.00** | 0 | 0 | 5.71 | 52.85 |
|  | Paste spot | 0 | 0 | 0.83 | **82.12** | 0.56 | 1.12 | 16.48 |
|  | Color difference | 0 | 0 | 0 | 0 | **100** | 0 | 0 |
|  | Dirty cells | 0 | 0.54 | 0.27 | 2.73 | 0 | **86.33** | 10.11 |
|  | Right | 0.32 | 0.96 | 0 | 0 | 0 | 0.64 | **98.06** |
| MS Solar Cell CNN | Thick line | **76.39** | 0 | 0 | 1.39 | 0 | 1.39 | 20.83 |
|  | Broken gate | 0 | **80.45** | 0 | 0 | 0 | 0.37 | 18.79 |
|  | scratches | 0 | 0 | **48.57** | 0 | 0 | 4.29 | 47.14 |
|  | Paste spot | 0 | 0 | 0.56 | **82.12** | 0.28 | 1.12 | 15.92 |
|  | Color difference | 0 | 0 | 0 | 0 | **100** | 0 | 0 |
|  | Dirty cells | 0.54 | 0.54 | 0 | 3.00 | 0 | **87.16** | 8.74 |
|  | Right | 0.32 | 0.96 | 0 | 0 | 0 | 0.64 | **98.06** |

It can be seen from the experimental results that the detection of solar cell surface defects using LBP+HOG-SVM and Gabor-SVM is not very effective. It can be seen that typical machine learning methods like LBP or Gabor with SVM's precision are 10% lower than CNN methods, and the recall are also 8% lower. The reason is that the LBP+HOG and Gabor features are more pronounced for texture features of defects in uniform background. However, there are many surface texture features on the cell surface, including a large number of non-defect background texture features such as lattices and grid lines, which may interfere with training. For machine learning methods that extract features manually, it is too difficult to attempt to express all the surface defect features of solar cell using high level features. Moreover, the defects of the solar cell have the characteristics of random shape and complex background, which makes the traditional machine learning method that requires manual extraction of features to perform the detection method to be less adaptable.

It can be found for in Fig.7 and Table 11 that the Solar CNN model shows stronger ability to distinguish multiple defect features, which helps to effectively solve the complex problem of irregular surface of cell surface defects and the random shape and color features of surface defects. Furthermore, from the experimental results, the multi-spectral solar cell CNN model has a higher accuracy and adaptability to the defect detection problem of random shape and complicated background on the surface of solar cells. It means that the ability of multi-spectral solar cell CNN model to extract features of different spectra is enhanced, and the defect features extracted by the model are more distinguishable.

The training and detection computation time for the two models are shown in Table 12. The training time of the multi-spectral solar cell CNN model is 6771 seconds longer than solar CNN model. However, multi-spectral solar cell CNN can speed up training through multi-threading, the performance of the experimental platform will be further improved. The detection time of the two models was calculated by testing 100 images. The solar CNN model took 3.66s while the multi-spectral solar cell CNN model took 4.25s. The detection time of each picture is within 50 milliseconds, so both neural networks meet the requirements of real-time monitoring. It should be mentioned that the deep learning model can input multiple image tensors at the same time, while the traditional machine vision method needs to read and extract features cyclically, so the total processing time per hundred images is quite different.

Table 11 Different results of training method

| Training methods | precision（%） | recall（%） | F-measure |
|---|---|---|---|
| LBP+HOG-SVM | 79.26 | 89.59 | 0.84 |
| Gabor-SVM | 74.55 | 89.26 | 0.81 |
| Solar cell CNN | 87.30 | 97.05 | 0.92 |
| MS-CNN | 88.41 | 98.40 | 0.94 |

Table 12 Training and detection time

| | Training time (s) | Detecting time (100 images) |
|---|---|---|
| Solar cell CNN | 4869 | 3.66 |
| multispectral-CNN | 11640 | 4.25 |
| LBP+HOG-SVM | 9785 | 42.20 |
| Gabor-SVM | 9670 | 35.70 |

## 5 Conclusion

Aiming at the wide variety of surface defects, various shapes, and severe background interference, the multi-spectral convolutional neural network model is proposed in this paper. Experimental results show that multi-spectral solar cell CNN model enhances the ability to extract multiple spectral information features, improves the ability to separate defects and background features, and improves the detection rate of most defects. The detection rate of non-defective pictures increased by about 1.4%. Therefore, the multi-spectral solar cell CNN model has higher accuracy and stronger adaptability in the detection of solar cell surface defects.

Although the multispectral convolutional neural network improves the detection results overall, the inadequacies can also be seen from the experimental data. In the experimental results, multi-spectral solar cell CNN models have relatively low detection rates for defects such as broken gates and scratches, and high detection rates for thick lines, dirty cell, paste spot, color difference. Analyzing the characteristics of these defects, it is found that the area of the broken gates and scratches with low detection efficiency is small and linear. However, the areas of defects such as color difference and dirty cell are relatively large and all had large area defects. The experimental results show that the multi-spectral solar cell CNN has weaker feature extraction ability for small area defects and linear defect defects.

The time consumed by multi-spectral solar cell CNN to detect 100 images is 3.66s, which only consumes 0.59s more than the 4.25s of the original CNN model. It still meets the processing needs of real-time detection. Subsequent preparations will further optimize the model's ability to identify linear and small-area defects, increase the training speed in a multi-threaded training mode, and test the model's ability to recognize more types of surface defects. In the future research, we are going to use the deep learning model to display the features extracted by the convolutional neural network to achieve accurate detection of defect locations. And by using the updated model, like VGG or ResNet as the base network of the RGB model, we will furtherly reduce the training time and over-fitting of the network.

ACKNOWLEDGEMENT

This work is supported in part by National Natural Science Foundation (NNSF) of China under Grant 61873315, Natural Science Foundation of Hebei Province under Grant F2018202078, Science and Technology Program of Hebei Province under Grant 17211804D, Hebei Province Outstanding Youth Science Foundation F2017202062 and Young Talents Project in Hebei Province under Grant 210003